\documentclass{article} 
\usepackage{iclr2020_conference,times}

\usepackage{amsmath,amsfonts,bm}

\def\eqref#1{equation~\ref{#1}}

\def\1{\bm{1}}

\DeclareMathAlphabet{\mathsfit}{\encodingdefault}{\sfdefault}{m}{sl}
\SetMathAlphabet{\mathsfit}{bold}{\encodingdefault}{\sfdefault}{bx}{n}

\usepackage{hyperref}
\usepackage{url}

\title{BERT Fine-tuning For \\ Arabic Text Summarization}

\author{Khalid N.~Elmadani \thanks{ These authors contributed equally.}~, Mukhtar Elgezouli \footnotemark[1]~ \& Anas Showk \\
Department of Electrical and Electronics Engineering\\
P. O. Box: 321 Khartoum, Sudan\\
\texttt{\{khalidnabigh,mukhtaralgezoli\}@gmail.com, anas.showk@uofk.edu}
}

\iclrfinalcopy 
\begin{document}

\maketitle

\begin{abstract}
Fine-tuning a pretrained BERT model is the state of the art method for extractive/abstractive text summarization, in this paper we showcase how this fine-tuning method can be applied to the Arabic language to both construct the first documented model for abstractive Arabic text summarization and show its performance in Arabic extractive summarization. Our model works with multilingual BERT (as Arabic language does not have a pretrained BERT of its own). We show its performance in English corpus ﬁrst before applying it to Arabic corpora in both extractive and abstractive tasks.\footnote{Our code is available at \href{https://github.com/mukhtar-algezoli/Arabic_PreSumm}{https://github.com/mukhtar-algezoli/Arabic\_PreSumm}.}
\end{abstract}

\section{Introduction}

Arabic, one of six official languages of the United Nations, is the mother tongue of 300 million people, and the official language for 26 countries; nine of those are in Africa. Hence Arabic had a huge influence in mother Africa forming the culture and religious values in West Africa, consequently it’s safe to say Arabic is "the Latin of Africa".

The Arabic script is an alphabet written from right to left. There are two types of symbols in the Arabic script: letters and diacritics~\citep{habash2010introduction}. It has 28 letters and each letter’s shape changes based on its position, each character holds 10 possible diacritics and the syntax of each word in the sentence depends on its last letter’s diacritic. Unfortunately, the diacritics are usually absent in the texts of news articles and any online content. The main challenge in Arabic text summarization is in the ambiguity of the Arabic language itself; the meaning of a text depends heavily on the context.

Text summarization is to extract and generate the key information in a brief expression from long documents. Generally, there are two approaches for text summarization either extractive: involves extracting the relevant phrases from the document, then organizing them to form the summary, or abstractive: involves going through the hole document, then try to write summary in your own words.
Arabic text summarization works are few related to other languages and the research is focused on extractive approaches which are based on sentences scoring then selecting best ones as a summary. Three approaches are used for sentence scoring and selection~\citep{al2017automatic}: symbolic-based systems: model the discourse structures of text, numerical-based systems: assign numerical scores to words of sentences, which reflects their significance, and hybrid systems: combine both symbolic-based and numerical-based methods. Recently,~\cite{jaafar2018towards}
suggested hybrid approach to produce abstractive summaries based on extractive ones.

\section{Data-sets}

English is the golden standard for text summarization, strongly because of the vast number of well proposed benchmark data-sets containing a huge capacity of summarized articles both in extractive and abstractive schemes like CNN/Daily-Mail news highlights data-set~\citep{hermann2015teaching} (contains 287K news articles and associated highlights), another important data-set in English is XSum~\citep{narayan2018don} (contains 226,711 news articles accompanied with a one-sentence summary, answering the question“What is this article about?”), this type of rich corpus is what Arabic language lacks in automatic text summarization.

The lack of Arabic benchmark corpora makes evaluation for Arabic summarization more difficult. Without unified benchmark corpus, the results reported from existing model can only be a hint for overall performance comparison~\citep{al2017automatic}.
But recently there is a turnout to use some corpora like EASC (containing 153 Arabic articles and 765 human-generated extractive summaries of those articles) and KALIMAT a Multipurpose Arabic Corpus (containing 20,291 articles with their extractive summaries).

\section{Methodology}

We used the pretrained BERT~\citep{devlin2018bert} for both abstractive and extractive summarization. The encoder (\textsc{BertSum})~\citep{liu2019text} is pretrained BERT expanded by adding several [CLS] symbols for learning sentence representations and using interval segmentation embeddings to distinguish multiple sentences. For abstractive summarization task the decoder is 6-layered Transformers~\citep{vaswani2017attention} initialized randomly. This mismatching between encoder and decoder -the encoder was pretrained while decoder is not- may lead to unstable training, so~\cite{liu2019text} proposed a new fine-tuning schedule which adopts different optimizers for the encoder and the decoder (\textsc{BertSumAbs}). And, for extractive summarization task a sigmoid classifier was inserted on top of each [CLS] token in the encoder indicating whether the sentence should be included in the summary (\textsc{BertSumExt}).

This method of using pretrained BERT is perfect for our condition, because the pretrained model will compensate for the relatively small data-set we are using.
But how could this model be applicable for Arabic Language since BERT was trained on English documents? The answer is Multilingual BERT (M-BERT)~\citep{pires2019multilingual}. It’s similar to the normal BERT but trained on 104 languages. We trained \textsc{BertSumAbs} one time using BERT and M-BERT another time on the CNN data-set for 45,000 steps to compare the impact of using M-BERT instead of BERT, sense M-BERT supports Arabic.

Finally, we included a non-pretrained Transformer baseline for both extractive and abstractive tasks, in order to measure the effect of using pretrained M-BERT. Both \textsc{TransformerAbs} and \textsc{TransformerExt} encoders are 6-layered transformers, the rest of their architecture is the same as \textsc{BertSumAbs} and \textsc{BertSumExt} respectively.


 

\section{Results}

We've automatically evaluated the quality of the summary using ROUGE \citep{ROUGE}. Unigram and bi-gram overlap (ROUGE-1 and ROUGE-2) are reported as a means of evaluating informativeness and the longest common subsequence (ROUGE-L) as a means of evaluating fluency.

Table \ref{sample-table} demonstrates the first step towards Arabic text summarization; switching from monolingual BERT to multilingual BERT. The results show very similar performance as compared to BERT and M-BERT.

Table \ref{table2} presents our results on KALIMAT data-set. We conclude that pre-trained M-BERT leads to huge improvements in performance for relatively small data-sets in both extractive and abstractive summarization. It also reveals that extractive models would have higher performance for extractive data-sets than their corresponding abstractive ones.

\begin{table}[t]
\caption{ROUGE F1 results on the CNN test set}
\label{sample-table}
\begin{center}
\begin{tabular}{llll}
\multicolumn{1}{c}{\bf Model}  &\multicolumn{1}{c}{\bf R1}  &\multicolumn{1}{c}{\bf R2} &\multicolumn{1}{c}{\bf RL}
\\ \hline \\
BERT         &30.45 &11.62 &28.02 \\
M-BERT             &30.35 &11.33 &25.3 \\
\end{tabular}
\end{center}
\end{table}

\begin{table}[t]
\caption{ROUGE F1 results on KALIMAT test set}
\label{table2}
\begin{center}
\begin{tabular}{llll}
\multicolumn{1}{c}{\bf Model}  &\multicolumn{1}{c}{\bf R1}  &\multicolumn{1}{c}{\bf R2} &\multicolumn{1}{c}{\bf RL}
\\ \hline \\
\textsc{BertSumExt}         &42.02 &24.59 &41.99 \\
\textsc{TransformerExt}             &28.75 &14.80 &28.72 \\
\textsc{BertSumAbs}         &12.21 &4.36 &12.19 \\
\textsc{TransformerAbs}             &6.93 &1.78 &6.88 \\
\end{tabular}
\end{center}
\end{table}

\section{Conclusion}

In this paper, we showed how multilingual BERT could be applied to Arabic text summarization and how effective it could be in low resource situations. Research in Arabic NLP is still in its infancy compared to English; abstractive text summarization was not attempted before (at the time of this submission) so there is no metrics output that we can evaluate against.

\bibliography{iclr2020_conference}
\bibliographystyle{iclr2020_conference}

\appendix
\section{Data Pre-Processing}
KALIMAT is a multipurpose Arabic corpus used mainly for extractive summarization, in this section we will show how we prepared it for BERT.

In the raw data-set each category (culture, economy, local-news, international-news, religion, and sport) has its own file containing articles of each month in a txt file, the text on those txt files are in Latin-1 (ironically Latin-1 does not support Arabic), so we converted it to UTF-8 and put it conveniently in a CSV file, we then made a standalone (.Story) file for each article and its summary, with the summary formulated as highlights at the end of the file.

At this stage we used the preprocessing suggested by \cite{liu2019text} but with some changes to make it work with Arabic:

\begin{enumerate}
    \item In the Standford corenlp file we replaced the stanford-corenlp-3.9.2 with stanford-arabic-corenlp-2018–10–05-models and used its path as the path for Stanford corenlp tokenizer.
    \item We then used the sentence splitting and tokenization as in the paper, which split the articles and summaries into sentences (list of vectors) put into a JSON file.
    \item Lastly, we tokenized the vectors using BERT vocabulary (multilingual BERT model) and formatted it to Pytorch files (.pt).
\end{enumerate}

At the end we got 8 Pytorch files each with 2000 entries, each entry containing:

\begin{enumerate}
    \item src\_txt and src: those the original articles and their tokenized counterparts (tokenized using multilingual BERT tokenizer).
    \item tgt\_txt and tgt: those the original summaries and their tokenized counterparts (tokenized using multilingual BERT tokenizer).
\end{enumerate}

\end{document}